\documentclass[11pt]{article}
\usepackage{cite}
\usepackage[margin=1in]{geometry}
\usepackage{amsmath,amssymb,amsfonts}
\usepackage{bm}
\usepackage{algorithmic}
\usepackage{graphicx}
\usepackage{xcolor}
\usepackage{booktabs}
\usepackage{caption}
\usepackage{indentfirst}
\usepackage[colorlinks=true,citecolor=blue,linkcolor=black,urlcolor=blue]{hyperref}

\usepackage{titlesec}

\titleformat{\section}
  {\centering\large\bfseries}
  {\thesection}{1em}{}

\titleformat{\subsection}
  {\normalsize\bfseries}
  {\thesubsection}{1em}{}

\titleformat{\subsubsection}
  {\normalsize\bfseries}
  {\thesubsubsection}{1em}{}

\titlespacing*{\section}{0pt}{1.2em}{0.8em}
\titlespacing*{\subsection}{0pt}{1.0em}{0.6em}
\titlespacing*{\subsubsection}{0pt}{0.8em}{0.5em}

\title{Neural Factorization-based Bearing Fault Diagnosis}

\author{
Zhenhao Li \thanks{College of Computer and Information Science, Southwest University, Chongqing, China. (\texttt{3177406997@qq.com})}
\and
Xu Cheng\thanks{Corresponding author. College of Computer and Information Science, Southwest University, Chongqing, China. (\texttt{1182130874@qq.com})}
\and
Yi Zhou \thanks{College of Vehicle Engineering, Chongqing Industry and Trade Polytechnic, Chongqing, China. (\texttt{zzzzy0312@163.com})}
\and
\thanks{This research is supported by the National Key Research and Development Program of China under grant 2024YFF0908200, the Fundamental Research Funds for the Central Universities SWU-KR24005 and Chongqing Natural Science Foundation under grants CSTB2024TIAD-CYKJCXX0028 and CSTB2025TIAD-STX0032.}
}

\date{} 

\begin{document}

\maketitle

\begin{abstract}
This paper studies the key problems of bearing fault diagnosis of high-speed train. As the core component of the train operation system, the health of bearings is directly related to the safety of train operation. The traditional diagnostic methods are facing the challenge of insufficient diagnostic accuracy under complex conditions. To solve these problems, we propose a novel Neural Factorization-based Classification (NFC) framework for bearing fault diagnosis. It is built on two core idea: 1) Embedding vibration time series into multiple mode-wise latent feature vectors to capture diverse fault-related patterns; 2) Leveraging neural factorization principles to fuse these vectors into a unified vibration representation. \cite{tang2025neural,lei2020applications,zhang2019deep,tang2025auto}.This design enables effective mining of complex latent fault characteristics from raw time-series data. We further instantiate the framework with two models CP-NFC and Tucker-NFC based on CP and Tucker fusion schemes, respectively. Experimental results show that both models achieve superior diagnostic performance compared with traditional machine learning methods. The comparative analysis provides valuable empirical evidence and practical guidance for selecting effective diagnostic strategies in high-speed train bearing monitoring\cite{r58,r19,r3,r6}.
\end{abstract}

\noindent\textbf{Keywords:} Anomaly Detection, Classification, Tensor Neural Network, Neural factorization, CP Decomposition, Tucker decomposition

\section{Introduction}
Rolling bearing is an indispensable part of rotating machinery. Its operation health is very important to the safety, reliability and efficiency of industrial system, and it is the main source of mechanical failure\cite{r70}. In recent decades, machine learning algorithms such as support vector machine and random forest have been widely used because of their interpretability and effectiveness for artificial features\cite{r61}.
 
Although these traditional machine learning methods are very popular, they also have inherent limitations. Their performance largely depends on the quality of manually designed functions, which requires considerable domain expertise\cite{b15}. In addition, although models such as random forest are robust to over fitting to some extent, it is difficult to autonomously learn hierarchical and complex representations from the original high-dimensional vibration signals. This may lead to poor diagnostic accuracy, especially when dealing with subtle failure modes or high noise industrial environments.

To solve these problems, this paper proposes a classification framework based on neural tensor decomposition. We propose two innovative large-scale models: neural CP decomposition model and neural Tucker decomposition model\cite{r63}. In this method, the tensor decomposition theory is introduced into the neural network structure, and the high-dimensional weight matrix is effectively decomposed into compact core tensor and factor matrix, which greatly reduces the number of trainable parameters. This not only alleviates the over fitting and improves the computational efficiency, but also inherently captures the multilinear and potential structures in the vibration data, which usually indicate specific failure modes.

The primary contributions of this work are summarized as follows:
\begin{enumerate}
    \item A novel neural factorization-based classification framework;
    \item It overcomes the dependence on artificial feature extraction on and can carry out powerful end-to-end learning directly from the original vibration signal;
    \item Two parameter-efficient bearing fault diagnosis models designed to scale effectively across datasets of varying sizes.
\end{enumerate}

Extensive experiments verify that the proposed models outperform traditional methods like Random Forest, while maintaining a highly efficient and compact architecture\cite{kapoor2023leakage,luo2018fast,tang2024temporal,breiman2001random}.

The remainder of the paper is organized as follows. Section \ref{sec:pre} presents the preliminaries. Section \ref{sec:method} introduces the proposed model. Section \ref{sec:empirical} reports the experimental results and analysis. Finally, Section \ref{sec:conclusion} concludes the paper and discusses future directions\cite{hearst1998support,chen2016xgboost,ayyadevara2018gradient,popescu2009multilayer}.
\section{Preliminaries}

\label{sec:pre}
In our proposed neural network architectures, we leverage both Canonical Polyadic (CP) and Tucker decompositions as powerful dimensionality reduction techniques \cite{r20,r55,r11,r38}. CP decomposition realizes feature fusion by Hadamard product of factor matrix, decomposes the original vibration signal into weighted combinations of multiple low rank components, and captures the core features of the signal in a compact form. Tucker decomposition uses double projection path and outer product operation to explicitly model the high-order interaction between different frequency band features through the core tensor, so as to realize the deep mining of complex fault modes in vibration signals. The two decomposition methods can significantly reduce the model parameters while retaining the key diagnostic information\cite{lavalley2008logistic,peng2022non,smith2015rolling,neupane2020bearing}.

\subsection{CP Decomposition}\label{AA}
The CP decomposition is a fundamental tensor factorization method that generalizes matrix decomposition techniques like the Singular Value Decomposition (SVD) to higher-dimensional data. It approximates an N-way tensor as a sum of rank-one tensors. For a third-order tensor $\mathcal{X} \in \mathbb{R}^{I \times J \times K}$, the CP decomposition is expressed as:
\begingroup

\begin{equation}
\mathcal{X} \approx \sum_{r=1}^{R} \mathbf{a}_r \circ \mathbf{b}_r \circ \mathbf{c}_r
\end{equation}
\endgroup
where $\circ$ denotes the vector outer product, and $\mathbf{a}_r$, $\mathbf{b}_r$, $\mathbf{c}_r$ are the factor vectors corresponding to each mode for the $r$-th component\cite{b20,b8,r88,r7}.

\subsection{Neural Tucker factorization}
Neural Tucker factorization (NeuTucF) \cite{tang2025neural} is a latent factorization of tensor model based on a general neural network within the Tucker decomposition framework. It can be expressed as
\begin{equation}
    \hat{y}_{ijk} = \sum_{p=1}^P \sum_{q=1}^Q \sum_{r=1}^R \mathcal{G} \odot \mathcal{T}_{ijk} = \sum_{p=1}^P \sum_{q=1}^Q \sum_{r=1}^R g_{pqr} t_{pqr}^{(ijk)},
\end{equation}
where ${\mathcal{T}}_{ijk}$ is the Tucker interaction tensor. We apply this decomposition method to neural network models\cite{salman2024random,pisner2020support,das2021logistic,bentejac2021comparative}.
\section{Methodology}
\label{sec:method}
\begin{figure}[t]
    \centering
    \includegraphics[width=1\linewidth]{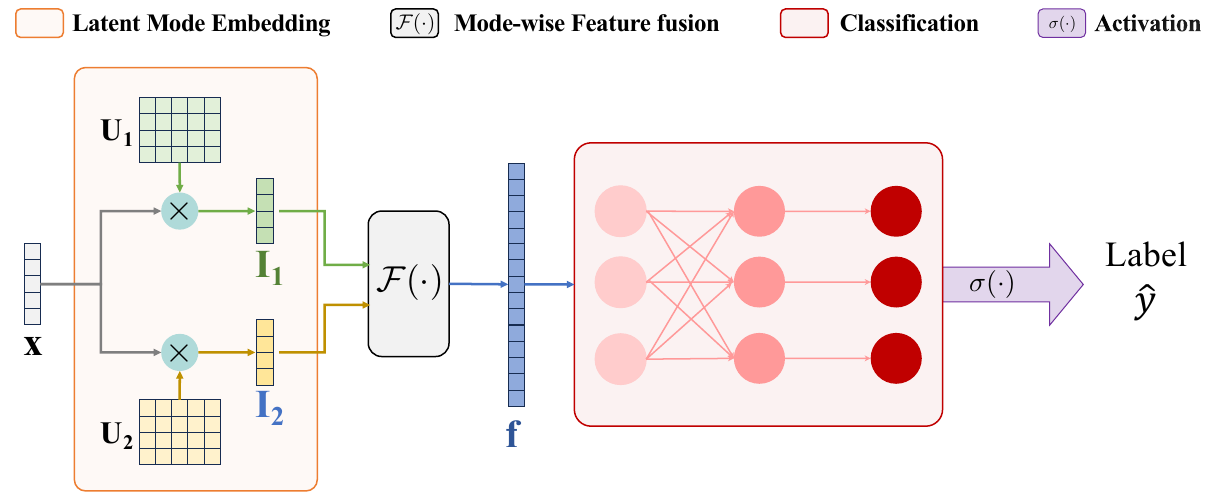}
    \caption{An illustration example of the proposed classification framework. The time series input is embedded into two different mode feature space.}
    \label{fig:mod}
\end{figure}

Taking inspiration from the principle of tensor decomposition and the NeuTucF framework, we propose a new tensor factorization-based classification framework for bearing fault analysis\cite{r17,r90,b16,r42}. It consists of three key components: latent mode embedding, multi-mode feature fusion and classification head. Figure \ref{fig:mod} shows the overall framework.

\subsection{Latent Mode Embedding}
To transform the input time series into a more expressive and compact representation for subsequent processes, we employ a latent feature embedding operation\cite{valkenborg2023support,hoang2019survey,li2002neural,mcinerny2003basic}. This layer embeds the time series segment into a lower-dimensional feature space while capturing different mode pattern of the data~\cite{r73,r4,r28}.
It is formulated as:
\begin{equation}
    \label{eq:latent_mode_embedding}
    \mathbf{I}_k = \mathcal{M}_k(\mathbf{x})=\mathbf{U}_k\times\mathbf{x},
\end{equation}
where $\mathcal{M}_k(\cdot)$ denotes the k-th mode feature embedding operation, $\mathbf{x}\in \mathbb{R}^{L}$ is the time series input with a length of $L$, $\mathbf{U}_k \in \mathbb{R}^{D_k \times L}$ is the $k$-th learnable embedding matrix, and $\mathbf{I}_k$ is the corresponding mode-wise embedding vector with the length of $D_k$\cite{r82}. In this paper, we we will focus the discussion on the mode-2 case. Figure \ref{fig:mod} illustrates the embedding process with a mode-2 example.

\subsection{Multi-mode Feature Fusion}
\begin{figure}[t]
    \centering\includegraphics[width=1\linewidth]{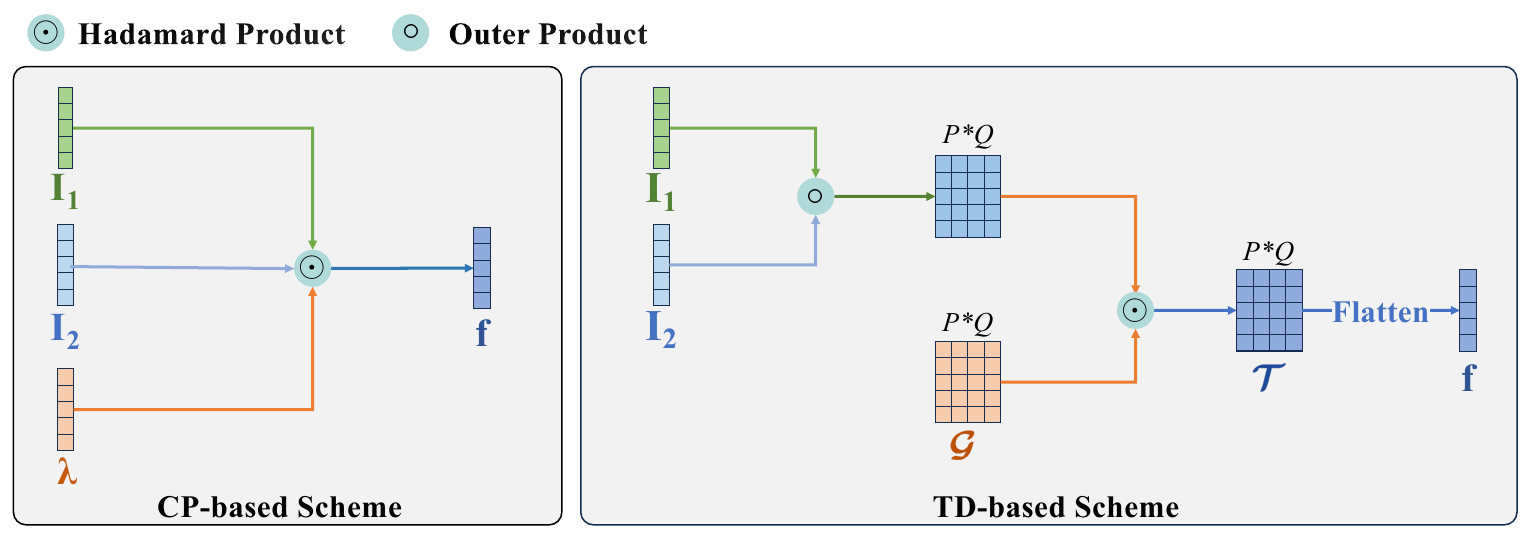}
    \caption{CP-based and TD-based Feature Fusion Schemes.}
    \label{fig:decomp}
\end{figure}

Multi-mode feature fusion module aggregates multiple mode embeddings into a single unified vector which captures high-order time series feature patterns, which is denotes as a single operation defined as 
\begin{equation}
    \mathbf{f}=\mathcal{F}(\mathbf{I}_1, \cdots, \mathbf{I}_k, \cdots, \mathbf{I_K})
\end{equation}
where $\mathbf{f}$ is a high-order vibration feature vector and $\mathcal{F}(\cdot)$ denotes a specific mode-wise fusion operation\cite{r33,b6,r68}.
In this study, we design and investigate two basic multi-mode feature fusion modules based on the principles of two typical tensor\cite{r44}. Figure \ref{fig:decomp} depicts the schemes of the two modules. Moreover, the framework can be easily extended to construct alternative fusion structures tailored to the characteristics of specific classification tasks~\cite{b10,r45,r40}.

\subsubsection{CP-based Scheme}
The CP-based feature fusion scheme is formulated as
\begin{equation}
    \mathbf{f} = \mathbf{\lambda} \odot (\mathbf{I}_1 \odot \mathbf{I}_2) 
\end{equation}
where $\mathbf{\lambda} \in \mathbb{R}^{R}$ is a learnable weight vector, $\odot$ denotes Hadamard product. In this scheme\cite{b4}, all the mode embedding vectors have the same size of $R$. Each component $f_r$ represents a distinct pattern extracted from the raw vibration signal, which is 
\begin{equation}
    f_r = \sum_{l=1}^L\lambda_rI_{ir}^{(1)}I_{ir}^{(2)}
\end{equation}

\subsubsection{Tucker-based Scheme}
For Tucker-based scheme, the projected features first undergo a bilinear fusion via an outer product to generate high-order feature interactions\cite{b5}. The level of feature interactions are weighted by a learnable core tensor to emphasize diagnostically relevant interactions\cite{r52}. The weighted fusion process for the mode-2 case is formulated as follows
\begin{equation}
    \boldsymbol{\mathcal{T}} = \boldsymbol{\mathcal{G}} \odot (\mathbf{I}_1 \circ \mathbf{I}_2),
\end{equation}   
where $\boldsymbol{\mathcal{T}} \in \mathbb{R}^{P \times Q} $ and $\mathcal{G} \in \mathbb{R}^{P \times Q}$ are the interaction tensor and the core tensor, respectively. $\circ$ and $\odot$ denote the outer product and Hadamard product, respectively\cite{r53}. Note that we slightly abuse the tensor notation here to highlight the model’s extendability to higher-order cases\cite{b3}.

The interaction tensor $\boldsymbol{\mathcal{T}}$ is then flattened into the vibration feature vector based on the following equation
\begin{equation}
    \mathbf{f} = \operatorname{Flatten}(\boldsymbol{\mathcal{T}})
\end{equation}
where the length of the feature vector $\mathbf{f}$ is $M=P\times Q$, and $\operatorname{Flatten}(\cdot)$ denotes the operation that unfolds a tensor into a vector while preserving the relative structural ordering of its elements~\cite{b9,r16,r8,r87}.

\subsection{Classification Head}
Finally, a deep fully connected network is employed as the classifier head to map the fused vibration feature vector to the output logits for bearing fault classification\cite{b7}. In this work, we adopt a hierarchical architecture composed of $N$ sequential fully connected layers, each consisting of batch normalization, a ReLU activation, and a dropout layer~\cite{r59}. The classifier is formally designed as
\begin{equation}
    \begin{cases}
        \mathbf{h}_{1} = \mathcal{H}_1(\mathbf{f}); \\
        \cdots \\
        \mathbf{h}_{N} = \mathcal{H}_N(\mathbf{h}_{N-1});
\end{cases}
\end{equation}
with the $n$-th fullly connected layer is 
\begin{equation}
    \mathcal{H}_n(\cdot) = \operatorname{Dropout}_n \Bigl(\operatorname{ReLU}_n \bigl(\operatorname{BatchNorm}_n(\cdot)\bigr)\Bigr).
\end{equation}

This article constructs a deep fully connected network as a classifier to complete the final fault classification task. This classifier adopts a hierarchical neural network architecture, consisting of multiple fully connected layers stacked in sequence\cite{r15}. Each fully connected layer is composed of a batch normalization layer, ReLU activation function, and Dropout layer. 

Through this progressive feature transformation and abstraction, the classifier can gradually map the high-dimensional feature representations extracted in the previous stage to a specific fault category space, thereby achieving accurate recognition and classification of different bearing fault modes\cite{r2}.
Through this progressive feature transformation and abstraction, the classifier can gradually map the high-dimensional feature representations extracted in the previous stage to a specific fault category space, enabling accurate identification of different bearing fault types from vibration signals~\cite{r60,r64,r65,r21}. 
\section{Empirical Study}
\label{sec:empirical}
\subsection{Dataset}
We conducted our experiments using the public CWRU Bearing 
Dataset\footnote{https://engineering.case.edu/bearingdatacenter}. Following prior studies\cite{hoang2019survey} and our own empirical analysis, we focused on four categories of drive-end (DE) vibration signals: Inner, Outer, Ball fault, and Normal\cite{neupane2020bearing}. Since DE faults primarily affect DE sensor measurements, only the DE time-series data were used for evaluation.

The raw vibration signals were segmented into fixed-length windows of 1024 sampling points with a 50\% overlap.\cite{r77} We adopt a stratified sampling method, retaining 20\% of the data for testing and the rest as training,while ensuring a balanced distribution of fault types\cite{r23}. All segments were then standardized to mitigate scale differences and enhance training stability~\cite{r39,r5,r10}.

\subsection{Models}
To establish a robust performance comparison, we carefully selected six widely used traditional machine learning modelsto represent different algorithm methods, including ensemble methods, linear models, gradient boosting, and neural networks. To ensure comprehensive coverage of different learning paradigms, the configuration details of each model are as follows~\cite{r10,r48,r30,r76}.
\begin{enumerate}
    \item M1: Random Forest (RF) with 200 estimators, maximum depth of 20, minimum samples split of 5, and minimum samples leaf of 2\cite{salman2024random};
    \item M2: Support Vector Machine (SVM) with Radial Basis Function kernel, regularization parameter $C$=1.0, and automatically scaled gamma parameter\cite{valkenborg2023support};
    \item M3: Extreme Gradient Boosting with 200 estimators, maximum depth of 10, and learning rate of 0.1\cite{chen2016xgboost};
    \item M4: Multi-layer Perceptron (MLP) with hidden layer architecture (128, 64, 32), Rectified Linear Unit activation function, and adaptive learning rate\cite{popescu2009multilayer};
    \item M5: Gradient Boosting Machine (GBM) with 200 estimators, learning rate of 0.1, and maximum depth of 5\cite{bentejac2021comparative};
    \item M6: Logistic Regression (LR) with L2 regularization, limited-memory BFGS optimization algorithm, and multinomial loss function\cite{das2021logistic};
    \item M7: Our  Neural CP Factorization  model;
    \item M8: Our  Neural Tucker Factorization  model\cite{tang2025neural};
\end{enumerate}

\subsection{Experiment Setting}
The experimental evaluation was carried out on a computing system equipped with a 2.50-GHz 13th Gen Intel Core i5-13400F processor, an NVIDIA GeForce RTX 3050 GPU, and 32 GB of system memory, leveraging Python 3.9 and the PyTorch 2.0.1 framework with CUDA 11.8 support. We adopt the widely used accuracy, precision, recall, and F1 score as our evaluation metrics for all the models in training.

Traditional models M1–-M6 rely on manual feature engineering to construct a set of interpretable features\cite{r74}. Accordingly, We carefully selected 12 time domain features, which are used to train M1–-M6 for bearing fault classification~\cite{r9}. In contrast, the proposed CP-NFC and Tucker-NFC models employ tensor algebra to perform end-to-end representation learning directly from raw bearing signals via tensor neural network architectures~\cite{r69}.

For a fair comparison, all models M1–-M8 follow the same data preprocessing pipeline and are evaluated under identical experimental conditions\cite{r56}. For training, we choose the Adam optimizer with an initial learning rate set to 1$e$-4, and apply a ReduceLROnPlateau scheduler~\cite{r22}. The batch size for model training is set to 32, and the model is trained for 100 epochs\cite{r57}. The hyperparameters of each baseline model are carefully selected based on prior knowledge to ensure strong and reliable performance·\cite{r84}.

\subsection{Results and Discussion}
Analysis of the experimental results shows that the tensor neural network models offer significant advantages in the CWRU bearing fault diagnosis task\cite{b17}. Overall, Tucker-NFC achieves the highest accuracy, precision, recall, and F1 score, clearly outperforming traditional machine learning methods\cite{r22}. Among the traditional models, XGBoost performs best, while logistic regression performs the worst, likely due to the limitations of its linear assumptions in handling complex fault modes\cite{r57}. In summary, our framework demonstrates strong potential for processing raw vibration signals and is particularly well-suited for complex industrial fault diagnosis scenarios\cite{r18}. Moreover, the end-to-end model design eliminates the need for manual feature extraction which hinders the application of other traditional classification models\cite{b19}.
\begin{table}[htbp]
  \centering
  \renewcommand{\arraystretch}{1.5} 
  \setlength{\tabcolsep}{4pt} 
  \caption{Benchmarking of Diagnostic Models on the CWRU Dataset. The best results are indicated in \textbf{bold}.}
  \label{tab:average_results}
  \begin{tabular}{lccccccc}
    \toprule
    \textbf{Model} & \textbf{Acc.} & \textbf{Prec.} & \textbf{Rec.} & \textbf{F1} & \textbf{Rank} & \textbf{Win/Loss} \\
    \midrule
    M1 & 0.8945 & 0.9011 & 0.8945 & 0.9133 & 3.0 & 0/4 \\
    M2 & 0.8278 & 0.8107 & 0.8278 & 0.8117 & 7.0 & 0/4  \\
    M3 & 0.8983 & 0.9020 & 0.8983 & 0.9000 & 2.0 & 0/4  \\
    M4 & 0.8963 & 0.8910 & 0.8963 & 0.8934 & 4.0 & 0/4 \\
    M5 & 0.8727 & 0.8764 & 0.8727 & 0.8731 & 5.0 & 0/4  \\
    M6 & 0.6720 & 0.6543 & 0.6720 & 0.6541 & 8.0 & 0/4  \\
    M7 & 0.8779 & 0.9169 & 0.8779 & 0.8931 & 6.0 & 0/4  \\
    M8 & \textbf{0.9080} & \textbf{0.9457} & \textbf{0.9080} & \textbf{0.9231} & \textbf{1.0} & \textbf{4/4}  \\ 
    \bottomrule
  \end{tabular}
\end{table}

\section{Conclusion}
\label{sec:conclusion}
In this paper, two new tensor neural network models, neural CP decomposition and neural Tucker decomposition, are proposed for end-to-end intelligent diagnosis of bearing faults\cite{r50}. The neural CP model realizes the efficient fusion of features through factorization, while the neural Tucker model explicitly models the high-order interaction between features through double projection and core tensor weighting mechanism\cite{r72}. Both methods break through the limitations of manual feature engineering in traditional fault diagnosis, and organically combine the mathematical interpretability of tensor decomposition with the powerful representation ability of deep learning\cite{b12}. Experiments on the CWRU bearing dataset show that the proposed method is significantly superior to the traditional machine learning method in the accuracy of fault classification, and provides a new and effective solution for rotating machinery fault diagnosis~\cite{r35}.

Future work can be carried out in the following directions: systematically benchmark different deep learning architectures, including convolutional neural networks, cyclic neural networks and attention mechanisms~\cite{r80,r14,r12}, which will provide empirical basis for the best modeling methods for different fault types. The in-depth study of multi-source signal fusion and model architecture optimization will promote the further development of adaptive learning framework in intelligent fault diagnosis\cite{b18}.

\bibliographystyle{unsrt}
\bibliography{reference_cheng}

\end{document}